\documentclass[runningheads]{llncs}

\usepackage[T1]{fontenc}
\usepackage{graphicx}
\usepackage{booktabs}
\usepackage{amsmath,amssymb}
\usepackage{hyperref}
\usepackage{tikz}

\begin{document}

\title{MPPI Planning with Gaussian Based Human Cost Function for Social Navigation}
\titlerunning{PGIF MPPI for Social Navigation}

\author{Chinmay Mundane}


\institute{VJTI, India\\
\email{cgmundane\_b21@pe.vjti.ac.in, chinmaymundane.123@gmail.com}}

\maketitle

\begin{tikzpicture}[remember picture,overlay]
\node[
    anchor=north,
    text width=0.88\paperwidth,
    align=center,
    font=\footnotesize
] at ([yshift=-8mm]current page.north) {%
\textbf{Preprint version.} This paper has been accepted for publication in the
\textit{Springer Proceedings in Advanced Robotics} as part of the proceedings
of the 19th International Workshop on Human-Friendly Robotics (HFR 2026),
Trento, Italy, July 16--17, 2026. The final authenticated version will be
available online at Springer Nature Link.%
};
\end{tikzpicture}

\begin{abstract}
Safe robot navigation in crowded spaces requires planning that accounts for where people will be, not only where they are now. Model Predictive Path Integral (MPPI) control is an effective
sampling-based planner, but many implementations encode humans as static point obstacles at their current positions, underestimating risk in dynamic scenes. We propose Predictive Gaussian Interaction Fields (PGIF), a spatiotemporal
cost formulation that propagates pedestrian predictions forward over the full planning horizon and encodes them as anisotropic Gaussian repulsive fields aligned with each pedestrian's direction of motion.
The forward spread of each field grows with the pedestrian's speed, creating a motion cone danger zone that penalises robot trajectories entering the pedestrian's path of travel more strongly than those approaching from behind. The formulation is closed-form and fully parallelisable across rollouts,
adding no measurable computational overhead.
Evaluated over 300 randomised crowd scenarios at three density levels,
PGIF-MPPI achieves a 0\% collision rate at every density level, compared
with up to 82\% for vanilla MPPI, while maintaining real-time planning performance. The implementation is publicly available at: \href{https://github.com/ChinmayMundane/PGIF_MPPI}{https://github.com/ChinmayMundane/PGIF\_MPPI}

\keywords{Mobile Robot Navigation, MPPI, Social Navigation, Human-Robot Interaction}
\end{abstract}

\section{Introduction}
Safe navigation among humans is one of the central open problems for autonomous mobile robots deployed in hospital corridors, shopping centres, pedestrian plazas, and logistics warehouses~\cite{kruse}. Human motion is inherently stochastic, socially structured, and temporally extended. For example, a pedestrian's position five seconds from now depends not only on
their current velocity but on their goals, group dynamics, and awareness of
the robot. A planner that cannot represent this temporal structure is limited in its ability to act safely and efficiently.

The consequences of ignoring future human motion range from minor inefficiency to dangerous failure.
The freezing robot problem~\cite{pete}, in which a robot confronted with a dense crowd halts because every candidate path appears to intersect
a human at the current instant, is one example of myopic planning. Conversely, a planner that aggressively pursues goal-reaching without anticipating pedestrian motion risks close encounters in crossing or merging scenarios where human and robot paths converge.
 
The MPC and MPPI literature already contains approaches that propagate pedestrian predictions through the planning horizon rather than treating pedestrians as static obstacles~\cite{dra,ref12}. However, these methods differ in how the predicted motion is incorporated into the planning cost. For example, DRA-MPPI~\cite{dra} estimates the probability of collision between the robot and predicted pedestrian distributions and penalizes trajectories according to the resulting risk. In contrast, PGIF directly encodes each predicted pedestrian as an anisotropic Gaussian interaction field aligned with the pedestrian's direction of motion. The field expands in the direction of travel as pedestrian speed increases, producing a direction-aware cost landscape that penalizes trajectories entering the pedestrian's future path more strongly than those approaching from the sides or behind. This representation provides continuous spatial guidance while remaining computationally efficient for MPPI optimization.
 
In this work, we ask a central question: how can a planner explicitly anticipate pedestrian trajectories without sacrificing the computational speed required for real-time navigation? Specifically, we seek a representation that penalises trajectories entering a pedestrian's predicted path more strongly than trajectories passing behind them, without introducing significant computational overhead. We address it as follows:
\begin{enumerate}
\item We formulate a spatiotemporal cost landscape over the MPPI planning
horizon by propagating pedestrian predictions forward in time and
encoding them as anisotropic Gaussian fields oriented along each
pedestrian's velocity direction. The forward spread grows with speed,
creating a motion cone danger zone. The formulation is closed form and
fully parallelisable over rollouts.
\item We benchmark PGIF MPPI against a vanilla, instantaneous-position MPPI baseline across 300
randomised corridor scenarios at three crowd density levels, reporting
collision rate, success rate, timeout rate, path length, and computation
time.
\item We characterise how and why PGIF MPPI becomes conservative at high crowd
densities, and discuss principled directions to recover goal reaching
without sacrificing safety.
\end{enumerate}

\section{Related Work}
\label{sec:related}

We organise prior work around the central question of this paper: how is
predicted human motion represented inside the planner's decision making,
and what is gained or lost by that representation. Classical reactive
methods (Sec.~\ref{sec:classical}) use little to no explicit prediction.
Learning-based methods (Sec.~\ref{sec:rl}) implicitly absorb prediction
into a learned policy at the cost of interpretability and sample
efficiency; sampling-based MPC methods (Sec.~\ref{sec:mppi}), including
ours, make the prediction explicit inside a cost or constraint, which is
interpretable and tunable but requires deciding how predictions are
spatially encoded; and trajectory prediction (Sec.~\ref{sec:pred}) and
Gaussian occupancy methods (Sec.~\ref{sec:gauss}) supply candidate
encodings for that representation. The gap we identify is that, within the
MPC/MPPI branch, existing predictive encodings are typically isotropic or
discrete-occupancy based, and do not jointly encode the directionality and
speed-dependent extent of pedestrian motion. PGIF is
designed to fill that gap.

\subsection{Classical Reactive Navigation}
\label{sec:classical}
 
The Social Force Model (SFM)~\cite{dirk} models pedestrian dynamics as a system of attractive and repulsive forces, enabling computationally efficient crowd simulation and robot navigation. When adapted for robot control, SFM suffers from oscillation in dense crowds and is fundamentally reactive so it cannot anticipate future pedestrian positions. Optimal Reciprocal Collision Avoidance (ORCA)~\cite{berg} computes velocity obstacles that guarantee collision free motion under the assumption of mutually cooperative agents. This assumption is frequently violated in real human-robot interactions, where humans do not actively yield to the robot~\cite{refsix}. Both approaches lack a long horizon planning mechanism and are susceptible to local minima in cluttered environments.
 
\subsection{Deep Reinforcement Learning for Social Navigation}
\label{sec:rl}
 
Learning-based approaches have demonstrated impressive social awareness, with models encoding pedestrian interactions implicitly through neural network representations~\cite{chen,eve}. These methods can capture complex group dynamics but inherit the well known limitations of deep RL like poor sample efficiency, sensitivity to reward shaping, difficulty in distributional generalisation, and absence of formal safety guarantees. Sim-to-real transfer also remains an open challenge, as neural policies trained on simplified pedestrian models often degrade in real world settings~\cite{de}. Unlike SFM/ORCA and the MPC-based approaches below, the prediction of human motion here is not represented explicitly. It is folded into the learned value function or policy, which makes the resulting avoidance behaviour difficult to inspect or certify.
 
\subsection{Sampling-Based MPC and MPPI}
\label{sec:mppi}
 
MPPI~\cite{will} was originally proposed for aggressive off-road driving
and has since been applied across a range of robotic platforms.
Extensions relevant to our work include learning cost functions from
demonstrations~\cite{ref10} and integrating neural dynamics models~\cite{ref11}.
In the context of social navigation, Williams et al.~\cite{ref12} incorporated pedestrian predictions into MPPI using discrete occupancy representations, while Trevisan et al.~\cite{dra} proposed a risk-aware MPPI framework that propagates obstacle trajectory predictions through the planning horizon and evaluates collision risk using Monte Carlo approximations. These approaches account for future pedestrian motion, but differ in how predicted occupancy is represented within the cost function. Our work focuses on this representation by introducing a continuous anisotropic Gaussian encoding of predicted pedestrian motion.
 
\subsection{Pedestrian Trajectory Prediction}
\label{sec:pred}
 
Predicting future pedestrian motion is an active research area~\cite{ref13}. Learning-based methods such as Social LSTM, Social GAN, and Trajectron++ can capture complex human interactions. However, for the short planning horizons typically used in MPPI, simple kinematic predictors are often sufficient and computationally efficient~\cite{ref14}. In this work, we use simple kinematic prediction and focus on how predicted pedestrian motion is encoded within the cost function. The proposed PGIF formulation is independent of the prediction method and can be combined with more advanced predictors.

\subsection{Gaussian Occupancy Representations}
\label{sec:gauss}
 
The continuous Gaussian encoding we employ is similar to Gaussian
Process occupancy maps~\cite{ref15}, but is computed in closed form without
kernel approximations, enabling online execution at planning frequencies.
Our anisotropic extension of orienting the kernel along the pedestrian's
velocity and scaling the forward spread with speed is, to our knowledge,
novel within this combination of closed-form evaluation and MPPI
integration, and is the principal methodological contribution of this work.

\section{Methodology}
\label{sec:method}

\subsection{System Overview}

The system comprises three tightly coupled modules operating at each planning cycle. First, a pedestrian state estimator that provides current positions and velocities of nearby humans. Then the PGIF generator that projects pedestrian predictions forward over the planning horizon into a time indexed repulsive cost field, and finally the MPPI controller that samples candidate trajectories, evaluates them against the composite cost function, and outputs control commands. All three modules execute sequentially within a single planning timestep, maintaining real-time feasibility.

\subsection{Robot Kinematic Model}
 
We model the robot with a kinematic unicycle model appropriate for
differential-drive platforms.
The state vector is
\begin{equation}
    \mathbf{x} = [x,\; y,\; \psi,\; v]^{\top},
\end{equation}
where $(x, y)$ is the robot position in the world frame, $\psi$ is the
heading angle, and $v$ is the longitudinal speed. We refer to this robot
state as the \emph{ego} state, and to quantities subscripted or labelled
``ego'' (e.g.\ $x^{(\tau,k)}_{\mathrm{ego}}$ in Section~\ref{sec:pgif}) as
belonging to the robot being planned for, as opposed to a pedestrian.
The control input is
\begin{equation}
    \mathbf{u} = [v_{\mathrm{ref}},\; \dot{\psi}_{\mathrm{ref}}]^{\top}.
\end{equation}
We integrate the kinematics with a semi-implicit (symplectic) Euler scheme:
the heading is updated first using the current yaw rate, and the new
position is then integrated using the \emph{updated} heading rather than
the heading at the start of the step. The discrete-time update over timestep $\Delta t$ is:
\begin{align}
    \psi_{t+1} &= \psi_t + \dot{\psi}_t\,\Delta t, \label{eq:psi}\\
    x_{t+1}    &= x_t + v_t \cos(\psi_{t+1})\,\Delta t, \label{eq:x}\\
    y_{t+1}    &= y_t + v_t \sin(\psi_{t+1})\,\Delta t. \label{eq:y}
\end{align}

\subsection{MPPI Control Framework}

Model Predictive Path Integral (MPPI) control~\cite{will} is a sampling-based
optimal control method. At each planning step, MPPI generates $K$ candidate
control sequences by perturbing a nominal control sequence
$\mathbf{U}=\{\mathbf{u}_0,\ldots,\mathbf{u}_{T-1}\}$. Each candidate is
propagated through the robot model and assigned a trajectory cost
$\mathcal{L}^{(k)}$.

The control sequence is updated using a weighted average of the sampled
perturbations:
\begin{equation}
    \mathbf{U}^{*} = \mathbf{U} + \sum_{k=1}^{K} w^{(k)}\,\boldsymbol{\epsilon}^{(k)},
\end{equation}
where the importance weights are
\begin{equation}
\label{eq:weights}
    w^{(k)} =
    \frac{
        \exp\!\left(-\dfrac{1}{\lambda}
        \bigl(\mathcal{L}^{(k)}-\mathcal{L}_{\min}\bigr)\right)
    }
    {
        \sum_{j=1}^{K}
        \exp\!\left(-\dfrac{1}{\lambda}
        \bigl(\mathcal{L}^{(j)}-\mathcal{L}_{\min}\bigr)\right)
    }.
\end{equation}

Lower-cost trajectories receive higher weights and therefore contribute more
strongly to the final control update. The parameter $\lambda$ controls the
selectivity of the weighting.

In our implementation, MPPI uses $K=512$ rollouts over a horizon of
$T=40$ steps with $\Delta t=0.1$\,s, corresponding to a 4\,s planning
horizon. Control inputs are constrained to
$v\in[10^{-6},2.0]$~m/s and
$\dot{\psi}\in[-1.5,1.5]$~rad/s.

\subsection{Pedestrian Prediction}

At planning time $t_0$, pedestrian $i$ has observed position
$\mathbf{h}_i^0=(h_{x,i}^0,h_{y,i}^0)$ and velocity
$\dot{\mathbf{h}}_i=(\dot h_{x,i},\dot h_{y,i})$.
Future pedestrian states are predicted at each horizon step
$\tau=0,\ldots,T-1$, corresponding to time
$t_0+\tau\Delta t$.

For pedestrians moving along straight paths, we use a constant-velocity
(CV) model:
\begin{equation}
h^{\tau}_{x,i}=h^{0}_{x,i}+\dot h_{x,i}\tau\Delta t,
\qquad
h^{\tau}_{y,i}=h^{0}_{y,i}+\dot h_{y,i}\tau\Delta t.
\end{equation}

For pedestrians following circular motion, we propagate the angular
position as
\begin{equation}
h^{\tau}_{x,i}=c_{x,i}+r_i\cos(\theta_i+\omega_i\tau\Delta t),
\qquad
h^{\tau}_{y,i}=c_{y,i}+r_i\sin(\theta_i+\omega_i\tau\Delta t),
\end{equation}
where $(c_{x,i},c_{y,i})$ is the orbit centre, $r_i$ the radius,
$\omega_i$ the angular velocity, and $\theta_i$ the current angle.

Predicted velocities are obtained analytically from the corresponding
motion model and are used to determine the pedestrian motion direction
for the PGIF formulation described in the next section. 
 
\subsection{Predictive Anisotropic Gaussian Interaction Fields (PGIF)}
\label{sec:pgif}
 
The central contribution of this work is the PGIF cost term.
Rather than encoding pedestrian $i$ as an isotropic Gaussian centred on
their predicted position, we orient the kernel along their direction of
motion and use \emph{different spreads} in the forward, backward, and
lateral directions.
This creates a motion cone danger zone: the cost is high in front of the
pedestrian (where their future path lies) and lower behind them.
 
Let $s_i^{\tau} = \|\dot{\mathbf{h}}_i^{\tau}\|$ be the
pedestrian's predicted speed at step $\tau$, and let
\begin{equation}
    \hat{\mathbf{u}}_i^{\tau} = \frac{\dot{{h}}_i^{\tau}}{s_i^{\tau}},
    \qquad
    \hat{\mathbf{n}}_i^{\tau} = \bigl(-\hat{u}_{y,i}^{\tau},\; \hat{u}_{x,i}^{\tau}\bigr)
\end{equation}
be the unit vector along and perpendicular to the pedestrian's motion. Then, the displacement from the predicted pedestrian position to the ego
(robot) position at rollout step $\tau$ is decomposed as:
\begin{align}
    d_{\parallel}^{(\tau,k,i)} &=
        \bigl(x^{(\tau,k)}_{\mathrm{ego}} - h^{\tau}_{x,i}\bigr)\hat{u}_{x,i}^{\tau}
      + \bigl(y^{(\tau,k)}_{\mathrm{ego}} - h^{\tau}_{y,i}\bigr)\hat{u}_{y,i}^{\tau}, \\
    d_{\perp}^{(\tau,k,i)}     &=
        \bigl(x^{(\tau,k)}_{\mathrm{ego}} - h^{\tau}_{x,i}\bigr)\hat{n}_{x,i}^{\tau}
      + \bigl(y^{(\tau,k)}_{\mathrm{ego}} - h^{\tau}_{y,i}\bigr)\hat{n}_{y,i}^{\tau},
\end{align}
where $\bigl(x^{(\tau,k)}_{\mathrm{ego}}, y^{(\tau,k)}_{\mathrm{ego}}\bigr)$
denotes the robot's own predicted position, $(x_{t},y_{t})$ from
Eqs.~\eqref{eq:x}--\eqref{eq:y}, at horizon step $\tau$ along MPPI rollout
$k$.
The anisotropic spread is defined as:
\begin{equation}
\label{eq:sigma}
\sigma_{\parallel}^{(\tau,i)} =
\begin{cases}
1.2 + 0.5\,s_i^{\tau}, & d_{\parallel}^{(\tau,k,i)} \geq 0, \\
0.5, & \text{otherwise},
\end{cases}
\qquad
\end{equation}

The lateral spread is fixed at
\[
\sigma_\perp = 0.6~\mathrm{m}.
\]
This configuration produces an anisotropic interaction field with a larger
influence ahead of the pedestrian than to the sides or behind. Although
different parameter values can be adopted for specific environments, the
qualitative behaviour of PGIF remains unchanged.

The forward spread $\sigma_{\parallel}$ grows linearly with the pedestrian's
speed, so faster pedestrians project a longer danger zone ahead of them.
The PGIF cost at planning step $\tau$ for rollout $k$ is then:
\begin{equation}
\label{eq:pgif}
    C_{\mathrm{human}}^{(\tau,k)} = \sum_{i=1}^{N}
        \exp\!\left(
            -\frac{1}{2}\left[
                \left(\frac{d_{\parallel}^{(\tau,k,i)}}{\sigma_{\parallel}^{(\tau,i)}}\right)^{\!2}
              + \left(\frac{d_{\perp}^{(\tau,k,i)}}{\sigma_{\perp}}\right)^{\!2}
            \right]
        \right).
\end{equation}
The total human cost for a rollout accumulates over the horizon:
\begin{equation}
\label{eq:total_human}
    \mathcal{C}_{\mathrm{human}}^{(k)} = \sum_{\tau=0}^{T-1} C_{\mathrm{human}}^{(\tau,k)}.
\end{equation}

\begin{figure*}[t]
    \centering
    \includegraphics[
        width=\textwidth,
        height=0.5\textheight,
        keepaspectratio
    ]{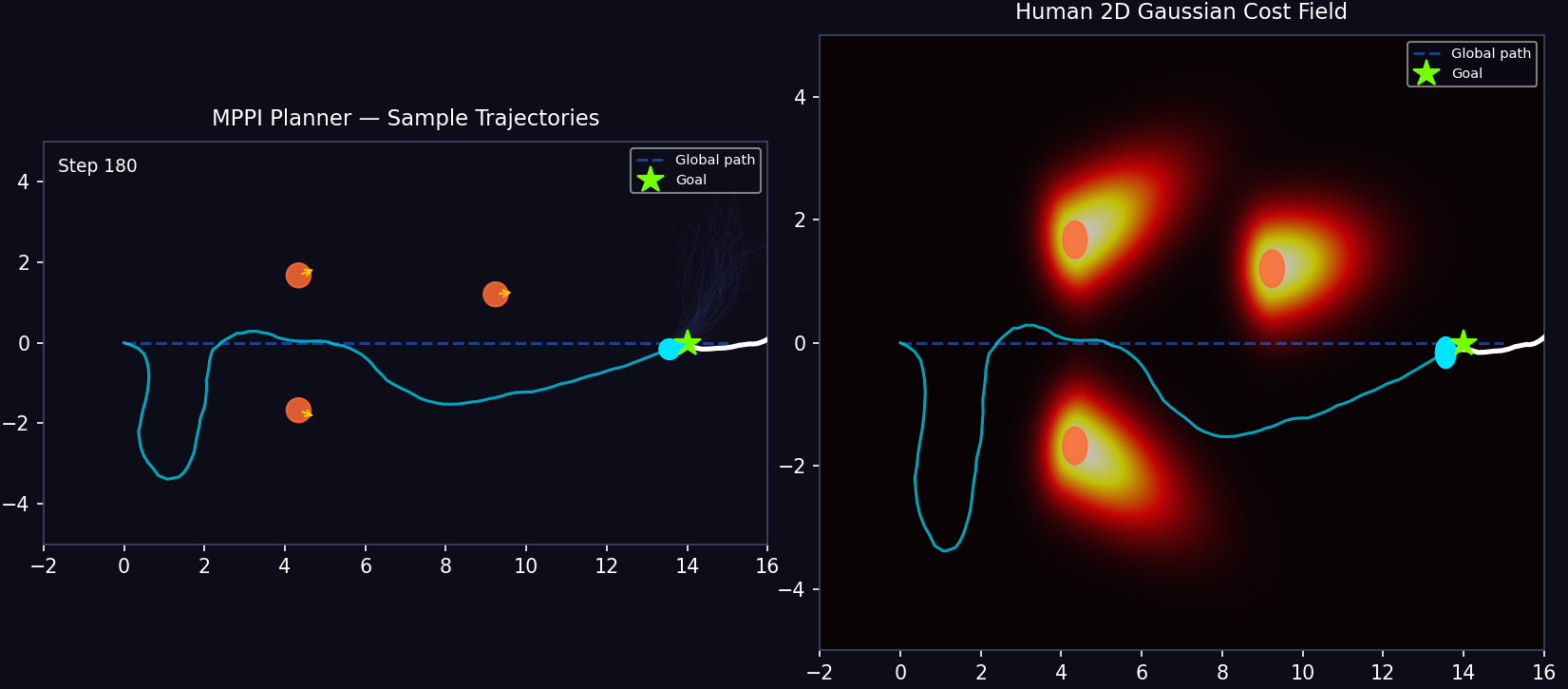}
    \caption{%
        \textbf{Left:} MPPI sample trajectories at a single planning
        step. The selected trajectory deviates above the direct path to
        avoid pedestrians (orange circles), each of which projects an
        anisotropic cost field elongated in their direction of travel.
        \textbf{Right:} The corresponding 2D anisotropic Gaussian cost
        field. High cost regions are elongated ahead of each
        pedestrian and narrower laterally and behind, reflecting the
        motion cone structure of the PGIF kernel. The planned trajectory
        passes through low cost corridors between these fields.%
    }
    \label{fig:cover}
\end{figure*}

 We note that PGIF is a soft cost rather than a hard safety constraint. Rollouts that pass through high-cost regions receive lower weights during the MPPI update but are not explicitly rejected. Consequently, the proposed method does not provide formal collision-avoidance guarantees, unlike  control-barrier function \cite{cbf} or chance-constrained \cite{chance} approaches. Instead, the results in Section~\ref{sec:results} demonstrate its practical effectiveness on the evaluated scenarios.

Fig.~\ref{fig:cover} illustrates the PGIF cost field together with MPPI rollouts at a single planning step. The anisotropic Gaussian produces a larger cost region in front of each pedestrian and smaller regions to the sides and behind. As pedestrians move, these fields are updated using their predicted positions, encouraging the planner to avoid likely future conflicts while allowing lower-cost paths around them.

The proposed kernel has three advantages. First, it provides a smooth cost that decreases continuously with distance, allowing MPPI to distinguish safe trajectories from near collisions. Second, its parameters have intuitive meanings and can be adjusted to suit different environments or navigation preferences. Third, the cost is evaluated in closed form and can be computed efficiently for all rollouts in parallel, adding negligible computational overhead.

\subsection{Complete Cost Function}
 
The full MPPI trajectory cost combines four terms:
\begin{equation}
\label{eq:cost}
\mathcal{L}^{(k)} =
w_{\mathrm{goal}}\,\mathcal{C}_{\mathrm{goal}}^{(k)} +
w_{\mathrm{term}}\,\mathcal{C}_{\mathrm{term}}^{(k)} +
w_{\mathrm{path}}\,\mathcal{C}_{\mathrm{path}}^{(k)} +
w_{\mathrm{human}}\,\mathcal{C}_{\mathrm{human}}^{(k)},
\end{equation}
where $\mathcal{C}_{\mathrm{goal}}$ is the minimum Euclidean distance from
any rollout state to the goal (per-step proximity reward);
$\mathcal{C}_{\mathrm{term}}$ is the squared Euclidean distance from the
\emph{terminal} rollout state to the goal;
$\mathcal{C}_{\mathrm{path}}$ is the cumulative lateral deviation from a
reference path provided by a global planner; and
$\mathcal{C}_{\mathrm{human}}$ is the PGIF cost defined in
\eqref{eq:pgif}-\eqref{eq:total_human}.
The high relative weight on $\mathcal{C}_{\mathrm{human}}$ reflects the
asymmetric importance of safety over efficiency in human-populated
environments. For vanilla MPPI, we use static obstacle cost instead of human cost term. All weights are fixed across all experimental scenarios.

\section{Experiments}
\label{sec:experiments}
All experiments are conducted in a custom 2D simulation environment implemented in Python. The environment represents a corridor scenario in which the robot navigates along the positive $x$-axis from the origin to a goal at $(14,0)$.
 
Pedestrian dynamics are modelled using two motion types commonly observed in crowd settings. Linear pedestrians move along straight crossing trajectories through the corridor with speeds uniformly sampled from $0.5$ to $1.2$~m/s. Circular pedestrians follow orbital paths around fixed centres near the corridor midline, with radii between $1.2$ and $2.0$~m and angular velocities in the range $0.4$ to $0.8$~rad/s. All pedestrian parameters are randomly initialised for each scenario. The robot operates with a maximum speed of $2.0$~m/s, and each episode runs for up to $200$ planning steps, corresponding to $20$s.
 
Three crowd density are defined based on the number and arrangement of pedestrians near the robot’s path. The first level contains one linear pedestrian with minimal interaction. The second includes three linear and one circular pedestrian, resulting in moderate interaction with frequent crossings. The third consists of five linear and two circular pedestrians, producing dense and highly dynamic interactions across the corridor. Each regime comprises 100 independent randomised trials. A trial is considered successful if the robot reaches within $0.5$~m of the goal without collision before the step limit. A timeout occurs if the step limit is reached without either success or collision.
 
Performance is evaluated using collision rate, success rate, timeout rate, average path length in metres, and average computation time per planning step in milliseconds. We compare two configurations: a vanilla, instantaneous-position MPPI baseline, where pedestrians are treated as static point obstacles at their current position, and the proposed method incorporating predictive anisotropic Gaussian fields over the planning horizon.
 
All experiments are executed on a system equipped with an NVIDIA RTX 3050 GPU. The MPPI controller is implemented in JAX using fully batched operations over $K$ rollouts via jit and vmap. Pedestrian states are obtained directly from the simulator, and implications for real-world deployment are discussed in Section~\ref{sec:discussion}.  There can be higher throughput from more optimised MPPI implementations \cite{sbmpc,mppi_numba}.
 
The MPPI configuration uses $K = 512$ rollouts over a horizon of $T = 40$ steps with $\Delta t = 0.1$s and temperature parameter $\lambda = 1000$. Cost weights are set to $w_{\mathrm{goal}} = 1000$, $w_{\mathrm{term}} = 1000$, $w_{\mathrm{path}} = 5000$, and $w_{\mathrm{human}} = 10^6$.

\section{Results}
\label{sec:results}

Table~\ref{tab:main} summarises performance across all difficulty levels. The proposed method achieves zero collisions across all 300 trials, while the baseline degrades significantly as crowd density increases.

\begin{table}[t]
\caption{Performance across difficulty levels (100 runs each).}
\label{tab:main}
\centering
\begin{tabular}{lccc}
\toprule
\textbf{Metric} & \textbf{Easy} & \textbf{Medium} & \textbf{Hard} \\
\midrule
\multicolumn{4}{l}{\textit{Vanilla MPPI}} \\
Success Rate       & 78.0\% & 29.0\% & 18.0\% \\
Collision Rate     & 22.0\% & 71.0\% & 82.0\% \\
Timeout Rate       & 0.0\%  & 0.0\%  & 0.0\%  \\
Avg.\ Path (m)     & 11.79  & 7.76   & 6.66   \\
Avg.\ Time (ms)    & 3.52   & 3.68   & 3.84   \\
\midrule
\multicolumn{4}{l}{\textit{Proposed MPPI (PGIF)}} \\
Success Rate       & \textbf{93.0\%} & \textbf{78.0\%} & \textbf{41.0\%} \\
Collision Rate     & \textbf{0.0\%}  & \textbf{0.0\%}  & \textbf{0.0\%}  \\
Timeout Rate       & 7.0\%  & 22.0\% & 59.0\% \\
Avg.\ Path (m)     & 14.43  & 16.87  & 18.75  \\
Avg.\ Time (ms)    & 3.06   & 3.23   & 3.42   \\
\bottomrule
\end{tabular}
\end{table}

The proposed approach is the only method that eliminates collisions entirely. In contrast, the baseline drops from 78\% success in sparse settings to 18\% in dense environments, with the majority of failures at high density resulting in collisions. This behaviour highlights a fundamental limitation of ignoring the temporal and directional structure of human motion where the planner selects trajectories that appear feasible at the current instant but intersect with pedestrian motion within the execution horizon.

\begin{figure}[t]
\centering
\includegraphics[width=\columnwidth]{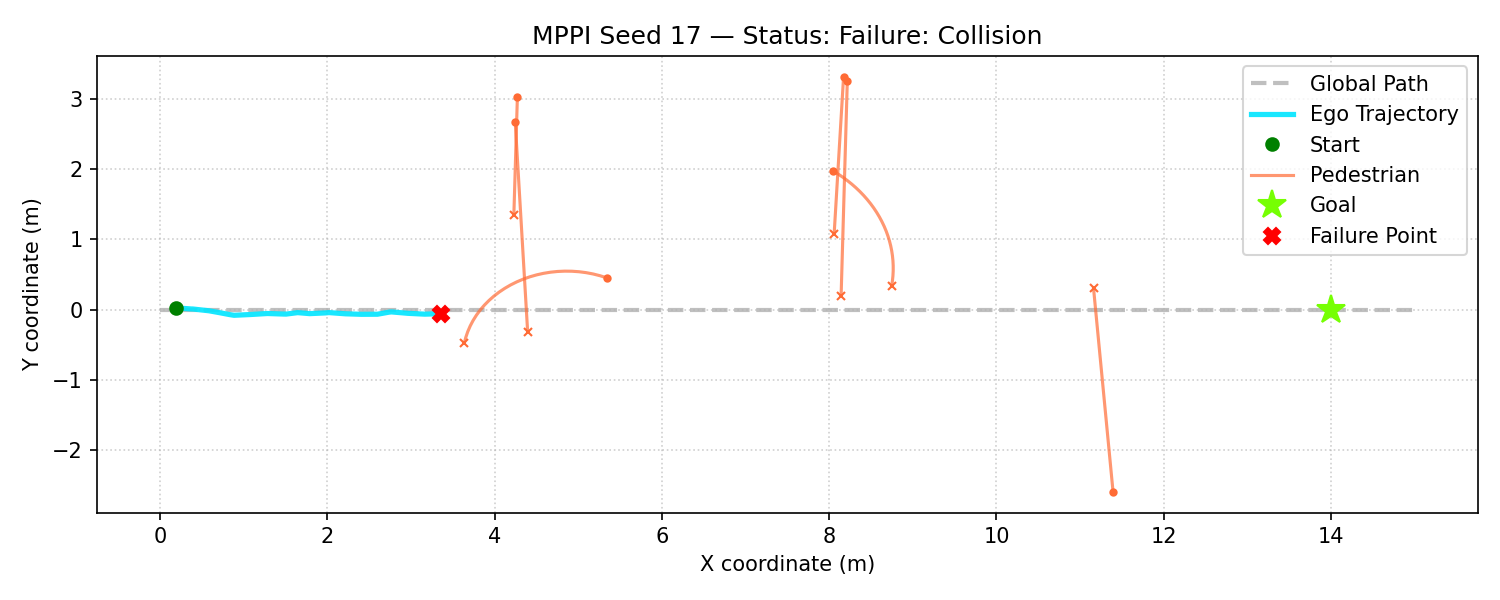}
\caption{
Vanilla MPPI failure case. The robot selects a trajectory that is collision-free with respect to current pedestrian positions, but pedestrians move into its path during execution, leading to collision.
}
\label{fig:vmppi}
\end{figure}

\begin{figure}[t]
\centering
\includegraphics[width=\columnwidth]{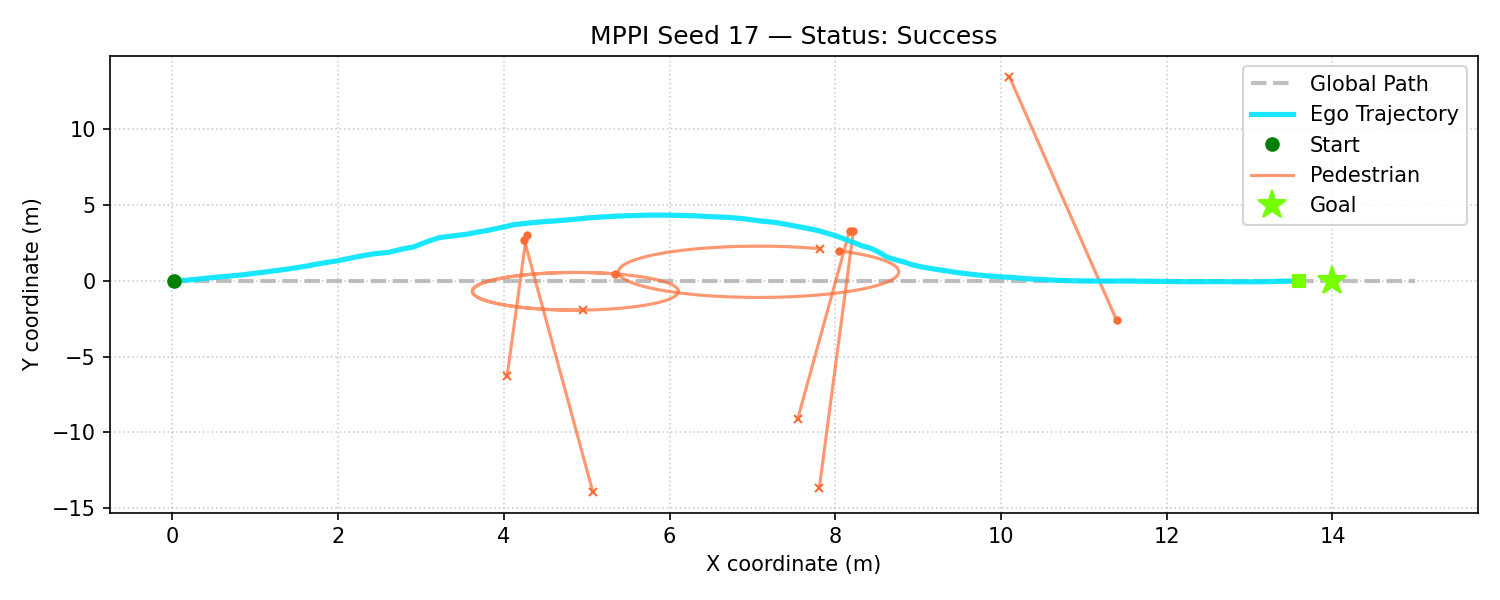}
\caption{
PGIF-MPPI behaviour in a comparable scenario. The planner deviates early from the nominal path, avoiding regions that are predicted to become occupied, resulting in a longer but collision-free trajectory.
}
\label{fig:gmmpi}
\end{figure}

This contrast is illustrated in Fig.~\ref{fig:vmppi} and Fig.~\ref{fig:gmmpi}. In the baseline case, the robot commits to a trajectory that is initially safe but intersects a pedestrian’s future path. With predictive cost encoding, the planner instead adjusts early, steering around regions of anticipated occupancy. This shift from reactive avoidance to anticipatory planning leads to safer behaviour at the expense of longer paths.

The increase in path length is consistent across all difficulty levels. For example, at high density the average path length increases from 14.43m to 18.75m. This reflects deliberate detours around predicted pedestrian motion. The shorter paths is observed in the baseline at higher densities  as many trajectories terminate early due to collision, reducing the measured distance travelled.

A limitation of the proposed method is the rise in timeout rate with increasing crowd density, from 7\% to 59\%. In dense scenarios, predicted anisotropic fields occupy a large portion of the workspace, leaving few low cost trajectories. As a result, the weighting mechanism favours trajectories with minimal forward progress, causing the robot to wait for the environment to clear. This behaviour is safe but suboptimal.

This form of conservatism differs from the classical freezing robot problem, which arises from the absence of collision free paths at the current time. Here, the issue stems from over weighting predicted future risk that has not yet materialised. Since it is tied to cost design rather than planner structure, it can be mitigated without altering the underlying MPPI framework.

Both methods satisfy real-time constraints. Average computation time remains well below the 4~ms. The predictive cost term introduces negligible overhead, with runtimes comparable to or slightly lower than the baseline. This is consistent with the $\mathcal{O}(K \cdot T \cdot N)$ complexity of the cost evaluation, which remains efficiently parallelisable on GPU hardware.

\section{Discussion}
\label{sec:discussion}

We employ two complementary prediction models: a constant-velocity model for pedestrians moving along straight trajectories, and exact circular kinematics for pedestrians following orbital motion. This choice is motivated by the relatively short planning horizon of 4s, within which simple kinematic predictions are typically sufficiently accurate. Moreover, since MPPI replans at every timestep, prediction errors are naturally corrected over time. While learned, goal conditioned predictors could improve performance in scenarios involving abrupt direction changes or complex group interactions, they would introduce additional model complexity and computational cost.

A key limitation of the proposed method is increased conservatism in high-density environments. As crowd density rises, the predicted anisotropic fields occupy a larger portion of the workspace, reducing the availability of low-cost trajectories. This leads to behaviours in which the robot delays forward motion in favour of maintaining safety. Several directions can address this limitation. One approach is to adapt the human cost weight dynamically, reducing $w_{\mathrm{human}}$ when no feasible low-cost rollout exists, thereby preventing excessive stagnation. Another approach is to scale the forward Gaussian spread with prediction uncertainty, allowing the influence of distant predictions to decay more gradually. A third direction is to incorporate intent-aware prediction, using estimates of pedestrian goals to avoid penalising regions that are likely to become unoccupied.

The current evaluation relies on ground-truth pedestrian states provided by the simulator. In real-world deployment, this assumption would need to be replaced with perception and tracking pipelines capable of estimating pedestrian positions and velocities reliably. Real environments also exhibit behaviours not captured in the simulation, including abrupt stops, group formations, and irregular motion patterns. Validation on a physical robotic platform operating in real pedestrian settings is therefore a necessary step toward practical deployment.

\section{Conclusion}
\label{sec:conclusion}

We introduced Predictive Gaussian Interaction Fields (PGIF), a principled
and lightweight extension to MPPI-based navigation that encodes
spatiotemporal pedestrian predictions as smooth, anisotropic repulsive cost
fields oriented along each pedestrian's direction of motion.
The forward spread of each field grows with the pedestrian's speed, creating
a motion-cone danger zone that more accurately represents where a moving
person will be compared to an isotropic kernel.
Evaluated across 300 randomised corridor scenarios, PGIF-MPPI achieves zero
collisions at all crowd density levels while maintaining real-time
computational performance.
The approach is modular, requires no learning, and integrates into any
existing MPPI system by adding a single cost term.
 
The primary remaining challenge is the conservatism efficiency trade-off in
dense, high-speed crowds.
Future work will focus on adaptive cost weighting, uncertainty-aware
Gaussian scaling, and validation with a physical robot in real pedestrian
environments.

\begin{credits}

\subsubsection{\discintname}
The author declares no competing interests.
\end{credits}

\bibliographystyle{splncs04}
\bibliography{references}

\end{document}